\documentclass{bmvc2k}

\title{Class-Balanced Distillation for Long-Tailed Visual Recognition}

\addauthor{Ahmet Iscen}{iscen@google.com}{1}
\addauthor{Andre Araujo}{andrearaujo@google.com}{1}
\addauthor{Boqing Gong}{bgong@google.com}{1}
\addauthor{Cordelia Schmid}{cordelias@google.com}{1}

\addinstitution{
 Google Research
}

\runninghead{Iscen \etal}{\CBD for Long-Tailed Visual Recognition}

\usepackage{times}
\usepackage{epsfig}
\usepackage{graphicx}
\usepackage{amsmath}
\usepackage{amssymb}
\usepackage{booktabs}

\usepackage{pifont}
\usepackage{multirow}
\usepackage{floatrow}

\usepackage{tikz}

\usetikzlibrary{arrows,shapes,calc,matrix,fit,backgrounds}
\usepackage{pgfplots}
\usepackage{pgfplotstable}
\pgfplotsset{compat=1.9}

\usepackage{xstring}

\usepgfplotslibrary{external}
\newcommand{\extdata}[1]{\input{#1}}
\IfBeginWith*{\jobname}{fig/extern/}{\finalcopy}{}

\newcommand{\leg}[1]{\addlegendentry{#1}}

\tikzset{every mark/.append style={solid}}
\pgfplotsset{%smooth,
	grid=both, width=\columnwidth, try min ticks=5,
	every axis/.append style={font=\scriptsize},
	every axis plot/.append style={thick,mark=none,mark size=1.2,tension=0.18},
	legend cell align=left, legend style={fill opacity=0.8},
}

\pgfplotsset{
	dash/.style={mark=o,dashed,opacity=0.7},
	dott/.style={mark=o,dotted,opacity=0.7},
}

\begin{document}

\maketitle

\newcommand{\nn}[1]{\ensuremath{\text{NN}_{#1}}\xspace}

\newcommand{\citemiss}{\alert{[??]}\xspace}

\newcommand{\supe}[1]{^{\mkern-2mu(#1)}}
\newcommand{\dime}[1]{(#1)}

\def\l1{\ensuremath{\ell_1}\xspace}
\def\l2{\ensuremath{\ell_2}\xspace}

\newcommand*\OK{\ding{51}}

\newenvironment{narrow}[1][1pt]
	{\setlength{\tabcolsep}{#1}}
	{\setlength{\tabcolsep}{6pt}}

%--------------------------------------------------------------------------------
%algorithm
\newcommand{\comment} [1]{{\color{orange} \Comment     #1}} % colored comment
\newcommand{\commentout}[1]{}
\newcommand{\prm}[1]{_{#1}}

%--------------------------------------------------------------------
\newcommand{\alert}[1]{{\color{red}{#1}}}
\newcommand{\head}[1]{{\smallskip\noindent\bf #1}}
\newcommand{\equ}[1]{(\ref{equ:#1})\xspace}

\newcommand{\red}[1]{{\color{red}{#1}}}
\newcommand{\blue}[1]{{\color{blue}{#1}}}
\newcommand{\green}[1]{{\color{green}{#1}}}
\newcommand{\gray}[1]{{\color{gray}{#1}}}

%--------------------------------------------------------------------

\newcommand{\tran}{^\top}
\newcommand{\mtran}{^{-\top}}
\newcommand{\zcol}{\mathbf{0}}
\newcommand{\zrow}{\zcol\tran}

\newcommand{\ind}{\mathbbm{1}}
\newcommand{\expect}{\mathbb{E}}
\newcommand{\nat}{\mathbb{N}}
\newcommand{\zahl}{\mathbb{Z}}
\newcommand{\real}{\mathbb{R}}
\newcommand{\proj}{\mathbb{P}}
\newcommand{\prob}{\mathbf{Pr}}

\newcommand{\mif}{\textrm{if }}
\newcommand{\other}{\textrm{otherwise}}
\newcommand{\minimize}{\textrm{minimize }}
\newcommand{\maximize}{\textrm{maximize }}
\newcommand{\st}{\textrm{subject to }}

\newcommand{\id}{\operatorname{id}}
\newcommand{\const}{\operatorname{const}}
\newcommand{\sgn}{\operatorname{sgn}}
\newcommand{\var}{\operatorname{Var}}
\newcommand{\mean}{\operatorname{mean}}
\newcommand{\trace}{\operatorname{tr}}
\newcommand{\diag}{\operatorname{diag}}
\newcommand{\vect}{\operatorname{vec}}
\newcommand{\cov}{\operatorname{cov}}

\newcommand{\softmax}{\operatorname{softmax}}
\newcommand{\clip}{\operatorname{clip}}

\newcommand{\defn}{\mathrel{:=}}
\newcommand{\peq}{\mathrel{+\!=}}
\newcommand{\meq}{\mathrel{-\!=}}

\newcommand{\floor}[1]{\left\lfloor{#1}\right\rfloor}
\newcommand{\ceil}[1]{\left\lceil{#1}\right\rceil}
\newcommand{\inner}[1]{\left\langle{#1}\right\rangle}
\newcommand{\norm}[1]{\left\|{#1}\right\|}
\newcommand{\frob}[1]{\norm{#1}_F}
\newcommand{\card}[1]{\left|{#1}\right|\xspace}
\newcommand{\diff}{\mathrm{d}}
\newcommand{\der}[3][]{\frac{d^{#1}#2}{d#3^{#1}}}
\newcommand{\pder}[3][]{\frac{\partial^{#1}{#2}}{\partial{#3^{#1}}}}
\newcommand{\ipder}[3][]{\partial^{#1}{#2}/\partial{#3^{#1}}}
\newcommand{\dder}[3]{\frac{\partial^2{#1}}{\partial{#2}\partial{#3}}}

\newcommand{\wb}[1]{\overline{#1}}
\newcommand{\wt}[1]{\widetilde{#1}}

\def\xssp{\hspace{1pt}}
\def\ssp{\hspace{3pt}}
\def\msp{\hspace{5pt}}
\def\lsp{\hspace{12pt}}

\newcommand{\cA}{\mathcal{A}}
\newcommand{\cB}{\mathcal{B}}
\newcommand{\cC}{\mathcal{C}}
\newcommand{\cD}{\mathcal{D}}
\newcommand{\cE}{\mathcal{E}}
\newcommand{\cF}{\mathcal{F}}
\newcommand{\cG}{\mathcal{G}}
\newcommand{\cH}{\mathcal{H}}
\newcommand{\cI}{\mathcal{I}}
\newcommand{\cJ}{\mathcal{J}}
\newcommand{\cK}{\mathcal{K}}
\newcommand{\cL}{\mathcal{L}}
\newcommand{\cM}{\mathcal{M}}
\newcommand{\cN}{\mathcal{N}}
\newcommand{\cO}{\mathcal{O}}
\newcommand{\cP}{\mathcal{P}}
\newcommand{\cQ}{\mathcal{Q}}
\newcommand{\cR}{\mathcal{R}}
\newcommand{\cS}{\mathcal{S}}
\newcommand{\cT}{\mathcal{T}}
\newcommand{\cU}{\mathcal{U}}
\newcommand{\cV}{\mathcal{V}}
\newcommand{\cW}{\mathcal{W}}
\newcommand{\cX}{\mathcal{X}}
\newcommand{\cY}{\mathcal{Y}}
\newcommand{\cZ}{\mathcal{Z}}

\newcommand{\vA}{\mathbf{A}}
\newcommand{\vB}{\mathbf{B}}
\newcommand{\vC}{\mathbf{C}}
\newcommand{\vD}{\mathbf{D}}
\newcommand{\vE}{\mathbf{E}}
\newcommand{\vF}{\mathbf{F}}
\newcommand{\vG}{\mathbf{G}}
\newcommand{\vH}{\mathbf{H}}
\newcommand{\vI}{\mathbf{I}}
\newcommand{\vJ}{\mathbf{J}}
\newcommand{\vK}{\mathbf{K}}
\newcommand{\vL}{\mathbf{L}}
\newcommand{\vM}{\mathbf{M}}
\newcommand{\vN}{\mathbf{N}}
\newcommand{\vO}{\mathbf{O}}
\newcommand{\vP}{\mathbf{P}}
\newcommand{\vQ}{\mathbf{Q}}
\newcommand{\vR}{\mathbf{R}}
\newcommand{\vS}{\mathbf{S}}
\newcommand{\vT}{\mathbf{T}}
\newcommand{\vU}{\mathbf{U}}
\newcommand{\vV}{\mathbf{V}}
\newcommand{\vW}{\mathbf{W}}
\newcommand{\vX}{\mathbf{X}}
\newcommand{\vY}{\mathbf{Y}}
\newcommand{\vZ}{\mathbf{Z}}

\newcommand{\va}{\mathbf{a}}
\newcommand{\vb}{\mathbf{b}}
\newcommand{\vc}{\mathbf{c}}
\newcommand{\vd}{\mathbf{d}}
\newcommand{\ve}{\mathbf{e}}
\newcommand{\vf}{\mathbf{f}}
\newcommand{\vg}{\mathbf{g}}
\newcommand{\vh}{\mathbf{h}}
\newcommand{\vi}{\mathbf{i}}
\newcommand{\vj}{\mathbf{j}}
\newcommand{\vk}{\mathbf{k}}
\newcommand{\vl}{\mathbf{l}}
\newcommand{\vm}{\mathbf{m}}
\newcommand{\vn}{\mathbf{n}}
\newcommand{\vo}{\mathbf{o}}
\newcommand{\vp}{\mathbf{p}}
\newcommand{\vq}{\mathbf{q}}
\newcommand{\vr}{\mathbf{r}}
\newcommand{\vt}{\mathbf{t}}
\newcommand{\vu}{\mathbf{u}}
\newcommand{\vv}{\mathbf{v}}
\newcommand{\vw}{\mathbf{w}}
\newcommand{\vx}{\mathbf{x}}
\newcommand{\vy}{\mathbf{y}}
\newcommand{\vz}{\mathbf{z}}

\newcommand{\vone}{\mathbf{1}}
\newcommand{\vzero}{\mathbf{0}}

\newcommand{\valpha}{{\boldsymbol{\alpha}}}
\newcommand{\vbeta}{{\boldsymbol{\beta}}}
\newcommand{\vgamma}{{\boldsymbol{\gamma}}}
\newcommand{\vdelta}{{\boldsymbol{\delta}}}
\newcommand{\vepsilon}{{\boldsymbol{\epsilon}}}
\newcommand{\vzeta}{{\boldsymbol{\zeta}}}
\newcommand{\veta}{{\boldsymbol{\eta}}}
\newcommand{\vtheta}{{\boldsymbol{\theta}}}
\newcommand{\viota}{{\boldsymbol{\iota}}}
\newcommand{\vkappa}{{\boldsymbol{\kappa}}}
\newcommand{\vlambda}{{\boldsymbol{\lambda}}}
\newcommand{\vmu}{{\boldsymbol{\mu}}}
\newcommand{\vnu}{{\boldsymbol{\nu}}}
\newcommand{\vxi}{{\boldsymbol{\xi}}}
\newcommand{\vomikron}{{\boldsymbol{\omikron}}}
\newcommand{\vpi}{{\boldsymbol{\pi}}}
\newcommand{\vrho}{{\boldsymbol{\rho}}}
\newcommand{\vsigma}{{\boldsymbol{\sigma}}}
\newcommand{\vtau}{{\boldsymbol{\tau}}}
\newcommand{\vupsilon}{{\boldsymbol{\upsilon}}}
\newcommand{\vphi}{{\boldsymbol{\phi}}}
\newcommand{\vchi}{{\boldsymbol{\chi}}}
\newcommand{\vpsi}{{\boldsymbol{\psi}}}
\newcommand{\vomega}{{\boldsymbol{\omega}}}

\newcommand{\rLambda}{\mathrm{\Lambda}}
\newcommand{\rSigma}{\mathrm{\Sigma}}

%--------------------------------------------------------------------
% Add a period to the end of an abbreviation unless there's one
% already, then \xspace.
% \makeatletter
% \DeclareRobustCommand\onedot{\futurelet\@let@token\@onedot}
% \def\@onedot{\ifx\@let@token.\else.\null\fi\xspace}
\def\onedot{.\xspace}
\def\eg{\emph{e.g}\onedot} \def\Eg{\emph{E.g}\onedot}
\def\ie{\emph{i.e}\onedot} \def\Ie{\emph{I.e}\onedot}
\def\cf{\emph{cf}\onedot} \def\Cf{\emph{C.f}\onedot}
\def\etc{\emph{etc}\onedot}
\def\vs{\emph{vs}\onedot}
\def\wrt{w.r.t\onedot} \def\dof{d.o.f\onedot}
\def\etal{\emph{et al}.}

\newcommand{\std}[1]{\tiny{$\pm$#1}}

\makeatother

\newfloatcommand{capbtabbox}{table}[][\FBwidth]

\definecolor{forestgreen}{RGB}{34,139,34}
\definecolor{red}{RGB}{255,0,0}

\newcommand{\accpos} [1]{\small\textbf{{\textcolor{forestgreen}{#1}}}}
\newcommand{\accneg} [1]{\small\textbf{{\textcolor{red}{#1}}}}

\newcommand{\acro}{\mathrm{{CBD}}}
\newcommand{\CBD}{$\acro$\xspace}
\newcommand{\CBDens}{$\acro_{\text{ENS}}$\xspace}
\newcommand{\CBDncm}{$\acro$.\textsc{ncm}\xspace}
\newcommand{\CBDnens}{$\acro_K$.\textsc{ncm}\xspace}

\newcommand{\instacro}{\mathrm{{Instance}}}
\newcommand{\crtacro}{\mathrm{{cRT}}}
\newcommand{\instcos}{$\instacro$.\textsc{cos}\xspace}
\newcommand{\crtcos}{$\crtacro$.\textsc{cos}\xspace}

\newcommand{\CBDinst}{$\acro$.\textsc{instance.cos}\xspace}
\newcommand{\CBDcrt}{$\acro$.\textsc{crt.cos}\xspace}
\newcommand{\CBDcens}{$\acro$.\textsc{crt.cos$_K$}\xspace}

\newcommand{\instlin}{\textsc{instance}\xspace}
\newcommand{\cbcos}{\textsc{cb.cos}\xspace}
\newcommand{\instncm}{\textsc{instance.ncm}\xspace}

%%%%%%%%% ABSTRACT
\begin{abstract}
Real-world imagery is often characterized by a significant imbalance of the number of images per class, leading to long-tailed distributions.
An effective and simple approach to long-tailed visual recognition is to  learn feature representations and a classifier separately, with instance and class-balanced sampling, respectively.
In this work, we introduce a new  framework, by making the key observation that a feature representation learned with instance sampling is far from optimal in a long-tailed setting.
Our main contribution is a new training method, referred to as Class-Balanced Distillation (\CBD), that leverages knowledge distillation to enhance feature representations.
\CBD allows the feature representation to evolve in the second training stage, guided by the teacher learned in the first stage. 
The second stage uses class-balanced sampling, in order to focus on under-represented classes.
This framework can naturally accommodate the usage of multiple teachers, unlocking the information from an ensemble of models to enhance recognition capabilities.
Our experiments show that the proposed technique consistently outperforms the state of the art on long-tailed recognition benchmarks such as ImageNet-LT, iNaturalist17 and iNaturalist18.\footnote{The code is available at \url{https://github.com/google-research/google-research/tree/master/class_balanced_distillation}}
\end{abstract}

%%%%%%%%% BODY TEXT
\section{Introduction}

Most of the modern computer vision techniques require large amounts of labeled training data in order to learn effective models, \eg,~for image classification~\cite{HZRS16,krizhevsky2012imagenet,SZ14}, object detection~\cite{HGD+17,RHG15}, image retrieval~\cite{radenovic2018revisiting,Noh_2017_ICCV,cao2020unifying} or segmentation~\cite{hu2018learning,chen2018encoder}.
Recently, much research has focused on learning with a smaller number of labels (\eg, few-shot learning \cite{DSH+18,GK18,SSZ17} or semi-supervised methods \cite{ITA+19,LA17,TV17}), or without any labels (\eg,~self-supervision \cite{caron2018deep,chen2020simple,grill2020bootstrap}).
While these works attempt at reducing the required annotations used for learning, they still tend to make the assumption that the training set is \emph{balanced}, meaning that there exists a similar number of examples per category.

\begin{figure*}[t]
\begin{center}
\includegraphics[width=0.90\textwidth]{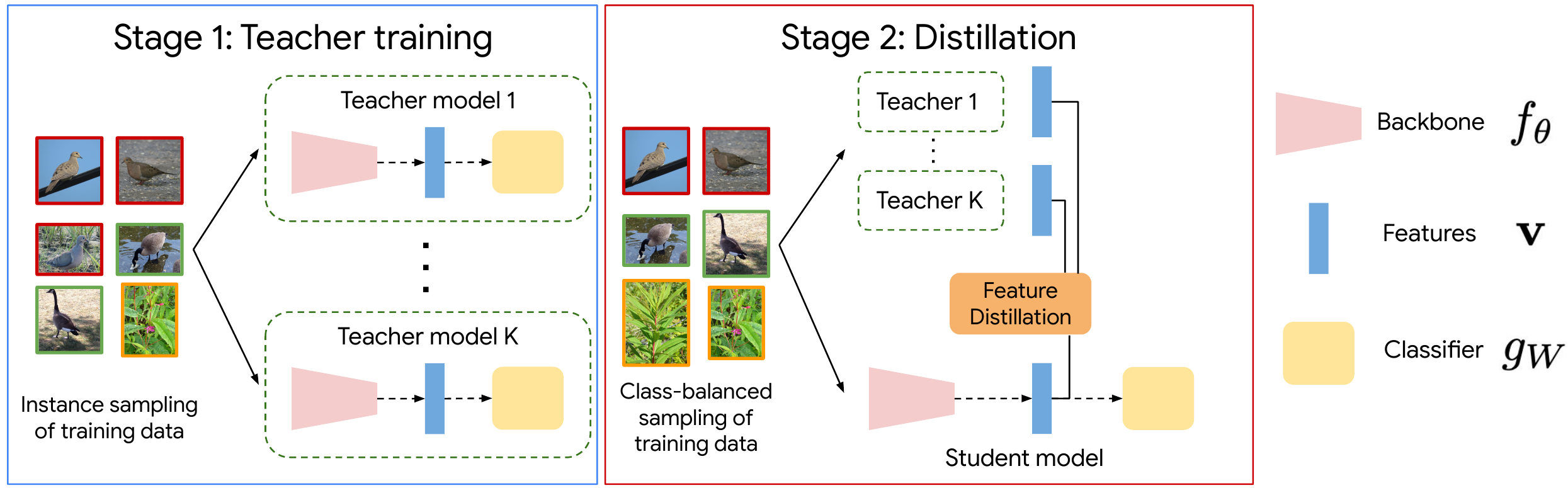}
\end{center}
\caption{
Overview of our Class-Balanced Distillation approach (\CBD). In the first stage, we learn one or multiple teacher models with instance sampling. In the second stage, we use class-balanced sampling to distill the features extracted by the teacher model(s) into a student model (right). The backbone is re-trained from scratch with feature distillation and a classification loss in the second stage. 
\label{fig:overview}
}
\end{figure*}

\emph{Long-tailed recognition} aims to address the real-world setting where a few of the labels are observed with very high frequency (\emph{head}), while most labels appear rarely (\emph{tail}), with a continuum in-between.
For example, in natural world datasets like iNaturalist~\cite{inaturalist}, some species are more abundant and easier to photograph than others; similarly, for datasets of human-made and natural landmarks~\cite{weyand2020google}, some are much more popular destinations than others.
This extreme imbalanced setting makes long-tailed visual recognition a challenging problem, where models often underfit the tail classes.
Early works tackle this challenge by different sampling strategies~\cite{cui2018large,LDAM} or re-weighting the loss function~\cite{CBLoss,Focalloss,khan2019striking}.

A very recent trend in this area is to (explicitly or implicitly) decouple the learning of the feature representation and the classifier into two stages~\cite{kang2019decoupling,Zhou_2020_CVPR,Jamal_2020_CVPR,chu2020feature,yang2020rethinking}. 
Typically, these methods first train a model with the imbalanced training data in the first stage, then apply additional operations, such as meta-learning instance-wise weights~\cite{Jamal_2020_CVPR} or augmenting the feature representations of under-represented classes~\cite{chu2020feature}, while they fine-tune the model in the second stage.
Kang \etal~\cite{kang2019decoupling} focus on the sampling strategies used in both stages and suggest that the feature representations are best learned with instance sampling (\ie, each image having the same probability of being sampled during training) in the first stage, while classifiers are best learned with class-balanced sampling (\ie, each class having the same probability of being sampled) in the second stage.

In our work, we propose a simple, flexible, and effective two-stage framework that 
makes a more aggressive decoupling of the two stages, allowing the second stage to learn a new feature extractor from scratch and the first stage to learn multiple, complementary models.
More specifically, we address two key observations that affect the existing approaches. 
The first observation is that the features learned by the instance sampling in previous works are far from optimal for a long-tailed dataset, which we demonstrate in Section~\ref{sec:exp}.
The second observation is that the class-balanced classifier learning improves \emph{tail} classes, but at the expense of penalizing \emph{head} classes.

We approach both shortcomings by class-balanced knowledge distillation~\cite{HVD15}, which allows the feature representations to continue evolving in the second stage and benefit from different sampling strategies.
Figure~\ref{fig:overview} illustrates the main components of  our method. 
We train an ensemble of teacher models with instance sampling in the first stage. In the second stage, we learn a student model with class-balanced sampling while distilling feature representations from the teachers. Compared with the training and fine-tuning strategy, our approach provides flexibility to the first stage, which can enhance the feature representation by ensembling, and a versatile distillation tool to the second stage, which essentially learns how to combine and evolve the features.
 
Our contributions are the following:
\begin{itemize}
\item A novel two-stage learning method, referred to as Class-Balanced Distillation (\CBD), which is suitable for long-tailed recognition datasets, simple to implement, and effective in combining the advantages of instance sampling and class-balanced sampling.
\item A feature distillation scheme for ensembling teachers, which efficiently combines feature representations of multiple teachers with different characteristics, including different data augmentations, to further improve its efficacy.
\item An extensive experimental evaluation of state-of-the-art long-tailed recognition benchmarks, demonstrating that our model outperforms prior arts substantially, with improvements for \emph{both} head and tail classes. 
\end{itemize}
\section{Related Work} \label{sec:related}

\head{Long-Tailed Recognition.}
The need for handling long-tailed datasets has emerged in many  applications, including but not limited to image classification~\cite{inaturalist2017,OLTR}, face recognition~\cite{Rangeloss,Unequal-Zhong_2019,Yin-feature-transfer,Cao_2020_CVPR}, object detection~\cite{Peng_2020_CVPR,Li_2020_CVPR}, instance segmentation~\cite{gupta2019lvis,wang2020devil,Hu_2020_CVPR}, and multi-label learning~\cite{wu2020distribution,Peng_2020_CVPR}. This work focuses on long-tailed image classification, but the proposed approach is generic and may benefit other applications. 

Some recent approaches decouple representation and classifier learning in deep long-tailed visual recognition~\cite{kang2019decoupling,Zhou_2020_CVPR,Jamal_2020_CVPR,chu2020feature,yang2020rethinking}. The representation learning stage often employs instance sampling, followed by different classifier learning methods. Kang~\etal \cite{kang2019decoupling} studied several normalization techniques for the linear classifier layer. Jamal~\etal~\cite{Jamal_2020_CVPR} proposed a meta-learning algorithm to re-weight both classes and instances. Zhou~\etal~\cite{Zhou_2020_CVPR} employed an annealing factor to transition the learning from representations to a classifier continuously. Chu~\etal~\cite{chu2020feature} augmented tail classes in the feature space. 
In contrast, we propose knowledge distillation~\cite{HVD15} as an efficient strategy for two-stage learning in long-tailed recognition, allowing the representation to evolve between different stages.
Besides, this enables learning from not just one, but an ensemble of teacher model representations.

Xiang \etal~\cite{xiang2020learning} have explored knowledge distillation in long-tailed classification for a different purpose from ours. 
The authors split the original long-tailed training set into a subset of more balanced training sets.
An \emph{expert} is learned for each subset, and distillation is used to fuse the experts into a single model.
In our work, we instead use the entire dataset for training the model and employ distillation to fuse the information from different teachers and sampling strategies into a single model.

Another line of research in long-tailed recognition is to promote the tail classes when training deep models.
These works include sampling the tail more frequently than the head~\cite{ren2020balanced}, re-weighting losses~\cite{CBLoss,ren2020balanced,Jamal_2020_CVPR}, balancing losses~\cite{Focalloss,LDAM,Tan_2020_CVPR,Liu_2020_CVPR}, and changing the momentum~\cite{tang2020long}.
Convolutional neural networks with memory modules may better represent the tail~\cite{OLTR,Zhu_2020_CVPR}, and one can also transfer knowledge from the head to the tail~\cite{Yin-feature-transfer,Kim_2020_CVPR,Modeltail-Wang}.
Wu~\etal~\cite{wu2020solving} introduced a taxonomic classifier to avoid making severe errors at the tail. These methods are orthogonal to ours, and they could complement each other. 

\head{Knowledge Distillation.}
Knowledge distillation~\cite{BCN06,HVD15} refers to transferring information from a \emph{teacher model} to a \emph{student model}.
It has been used in a variety of machine learning and computer vision tasks, such as image classification~\cite{HVD15}, object detection~\cite{CCYH+17}, semi-supervised learning~\cite{TV17} and few-shot learning~\cite{DSM19}.
Typically this involves making the output (logits) of student model similar to the teacher model.
In this work, we use a variant which transfers information directly at the feature level.
Feature distillation has been successfully used in other tasks, such as asymmetric metric learning~\cite{budnik2020asymmetric}.
It is also shown that feature distillation helps reduce catastrophic forgetting in incremental learning~\cite{hou2019learning, iscen2020memory} and domain expansion~\cite{jung2018less}.
In our work, we extend feature distillation to the case of multiple teacher models with different data augmentation and sampling.
\section{Method}
\label{sec:method}

\subsection{Classifier Training}
\label{sec:back}

\head{Problem Formulation.} We are given a set of $n$ instances (images) $X \defn \{x_1, \ldots, x_n\}$. Each image is labeled according to $Y \defn \{y_1, \ldots, y_n\}$ with $y_i \in C$, where $C \defn \{1,\dots,c\}$ is a label set for $c$ classes.
Let $C_j$ denote the subset of instances labeled as class $j$, and $n_j = |C_j|$ its cardinality.
In this paper, the training set follows a long-tailed distribution. 
Despite the training set imbalance, the goal is to accurately recognize all classes, so we use a balanced test set to evaluate the classifier. %(the number of test images for each class is the same).

\head{Model.} 
The learned model (typically a convolutional neural network) takes an input image and outputs class confidence scores. 
We denote the model by $\phi\prm{\theta,W}: \cX \rightarrow \real^c$.
It contains two components, corresponding to the learnable parameters $\theta$ and $W$, respectively:
1) a \emph{feature extractor}, mapping each instance $x_i$ to a descriptor $\vv_i \defn f\prm{\theta}(x_i) \in \real^d$;
2) a \emph{classifier}, typically consisting of a fully connected layer which output \emph{logits} $\vz_i \defn g\prm{W}(\vv_i) \in \real^c$, denoting the class confidence scores.

In this work, we model $g\prm{W}$ as a \emph{cosine classifier}~\cite{LZX+18,GK18}, where the feature descriptors and classifier weights are $\ell_2$-normalized before the prediction.
Its output becomes $\vz_i \defn \gamma \; \overline{W}^T \overline{\vv_i}$, where $\overline{\va}$ is the $\ell_2$-normalized version of $\va$, and $\gamma$ is a scaling hyper-parameter.
For simplicity, we omit the extra notation for $\ell_2$-normalization and refer to ${\vv_i}$ and $W$ as the $\ell_2$-normalized versions for the rest of this paper.

\head{Training.}
The model parameters $\theta$ and $W$ are typically learned by  minimizing the loss of the model's predictions over the training set $X$:
\begin{equation}
L( X, Y; \theta, W) \defn \sum_{i=1}^n \ell \left( \sigma(\vz_i), y_i\right),
\label{eq:loss}
\end{equation}
where $\vz_i = \phi\prm{\theta,W}(x_i)$ is the output of the model, $\sigma(.)$ is the softmax activation function, and $\ell(.)$ is the cross-entropy loss function.

\subsection{Sampling and Two-Stage Training}
\label{sec:baselines}

In the context of long-tailed problems, different sampling strategies have been used to adjust the data distribution at the training time.
We briefly review two sampling methods, which are utilized in this work. 

\head{Instance sampling} attributes each instance $x_i \in X$ with the same probability to a mini-batch. Hence,
the instances from the head classes are sampled more frequently than those from the tail classes due to the long-tailed nature of the dataset, making the model prone to underfitting tail classes.
Formally, let us denote by $p_j$ the probability of sampling an instance from class $j$.
Under instance sampling, $p_j = n_j / n$. 

\head{Class-balanced sampling} addresses the class imbalance by equalizing $p_j$ across classes.
Under this strategy, each class has the same probability of being selected, \ie, $p_j = 1/c$ for all $j = 1, \ldots, c$.
Even though this strategy balances the data distribution, it also under-utilizes the examples from the head classes.
Tail classes are sampled much more frequently compared to head classes.
As a result, the model tends to overfit the tail classes and exhibits sub-optimal  performance.

Two-stage approaches recently show improved performance for long-tailed recognition~\cite{kang2019decoupling,Zhou_2020_CVPR,Jamal_2020_CVPR,chu2020feature,yang2020rethinking}. 
We briefly review a few methods in this section; please see Section~\ref{sec:related} for a more thorough review. 

\head{Classifier Re-Training} (cRT) learns the two components of the model $\phi\prm{\theta,W}$ with different sampling strategies~\cite{kang2019decoupling}.
The feature extractor $f\prm{\theta}$ is first trained with instance sampling and then frozen, followed by learning the classifier $g\prm{W}$ with class-balanced sampling.
The authors argue that the first stage produces generalizable features, while the second stage makes the classifier less biased.

\head{Fine-tuning} 
trains the model $\phi\prm{\theta,W}$ with instance sampling in the first stage.
Then the the entire model $\phi\prm{\theta,W}$ is fine-tuned with class-balanced sampling, using a small learning rate for some number of epochs. 
The class-balanced sampling is vital for promoting the classifier's performance on the tail classes.
 
\head{Discussion.}
Instance sampling produces better feature representations compared to other sampling strategies~\cite{kang2019decoupling}.
However, the model's classifier is biased towards the head classes.
Two-stage methods leverage instance and class-balanced sampling separately to find the right balance between the two sampling strategies.
Classifier Re-Training learns the feature representations with instance and the classifier with class-balanced sampling, in this order~\cite{kang2019decoupling}.
While being simple and efficient, it has at least two shortcomings: (1) the feature representations tend to mostly focus on the head classes  due to the instance sampling in the first stage; (2) the second-stage, class-balanced classifier learning, could overcompensate tail classes, leading to reduced performance for the head classes.

\subsection{Class-Balanced Distillation (CBD)}
\label{sec:ltdistill}

To overcome the shortcomings in existing two-stage methods, we enhance the two-stage learning for long-tailed recognition by improving both (1) the feature representations for tail classes and (2) the classifier for head classes.
We leverage distillation~\cite{HVD15} to do so.
Figure~\ref{fig:overview} illustrates our overall approach.
In the first stage, we use instance sampling to train a teacher model $\widehat{\phi}\prm{\widehat{\theta},\widehat{W}}$.
In the second stage, we adopt class-balanced sampling and yet learn our student model $\phi\prm{\theta,W}$ from scratch by adding a feature distillation loss.

The feature distillation loss encourages the feature extractor $f\prm{\theta}$ of the student to heed the teacher's feature extractor.
It also amends the student's feature extractor to facilitate the classifier $g_W$. 
It reuses but does not fully inherit the first-stage's knowledge, leaving room for improvement with the class-balanced training.
The loss objective from Eq.~\eqref{eq:loss} becomes:
\begin{align}
L( X, Y; \theta, W) \defn & \sum_{i=1}^n (1-\alpha) \cdot \ell\left( \sigma(\vz_i), y_i\right) \nonumber \\
  & + \alpha \cdot \left( \beta \ell_F \left( \vv_i, \widehat{\vv_i}\right)\right), \label{eq:featdistill}
\end{align}
where $\widehat{\vv_i} = \widehat{f}\prm{\widehat{\theta}}(x_i)$ is the feature descriptor produced by the teacher model, and $\ell_F(\vv, \vx) = 1 - \text{cos}(\vv,\vx)$ tries to minimize the cosine distance between two feature descriptors. The hyper-parameter
$\alpha$ controls the amount of distillation compared to the cross entropy loss, and $\beta$ is a scaling parameter. 

\head{Feature-Level vs.\ Classifier-Level Distillations.}
Note that our objective function differs from the common knowledge distillation~\cite{BCN06,HVD15}, which is applied to to the classifier level rather than the feature level:
\begin{align}
L( X, Y; \theta, W) \defn & \sum_{i=1}^n (1-\alpha) \cdot \ell\left( \sigma(\vz_i), y_i\right)  \nonumber \\
  & + \alpha \cdot T^2 \cdot \ell \left( \sigma(\vz_i / T), \sigma(\widehat{\vz_i} / T )\right), \label{eq:distill}
\end{align}
where $\widehat{\vz_i}  = \widehat{\phi}\prm{\widehat{\theta},\widehat{W}}(x_i)$ is the teacher model's output, and $T$ is the temperature parameter used for distillation \cite{HVD15}. 

We experimentally show that the feature-level distillation is advantageous over the conventional classifier-level distillation.
In the context of long-tailed recognition, the teacher's classifier is highly biased towards the head classes.
By distilling only at the feature level (Eq.~\eqref{eq:featdistill}), we encourage the student to heed the teacher's feature extraction mechanism, not the classification function, to avoid learning a classifier that is significantly biased to the head.

\head{Distilling Ensemble of Teachers.}
Unlike the existing two-stage methods which learn a classifier (e.g., by cRT) or fine-tune the model, it is straightforward to use the proposed \CBD to further transfer knowledge from multiple teacher models.
The resulting student model, in this case, tends to have stronger regularization properties and reduced over-fitting~\cite{HVD15}.

To enable such capabilities, we train different teacher models with different characteristics.
More specifically, we train two types of teacher models with different data augmentations.
The \emph{Standard} model relies on standard data-augmentation transformations during training, such as random crop and flip.
The \emph{Data Augmentation} model uses additional data transformations, such as color jitter and Gaussian noise ($
\sigma=0.01$) in addition to random crop and flip.
When training multiple models of the same type, 
we start from different initial random seeds.
Different initial random seeds affect the initialization of the model parameters as well as the order of classes sampled during the training.
Regardless of the teacher model type, the \emph{standard} model is always used when training the student model in the second stage, according to our preliminary experiments.

Let $\widehat{\phi}^k\prm{\widehat{\theta}^k,\widehat{W}^k}$ denote the $k$-th teacher model.
When training the student model $\phi\prm{\theta,W}$ in the second stage, we combine the knowledge from multiple teachers with the following objective:
\begin{align}
L( X, Y; \theta, W) \defn & \sum_{i=1}^n (1-\alpha) \cdot \ell\left( \sigma(\vz_i), y_i\right)  \nonumber \\
  & + \alpha \cdot \left( \beta \ell_F \left( h(\vv_i), \widehat{\vV_i} \right)\right), \label{eq:ensembledistill}
\end{align}
where $\widehat{\vV_i} = [\widehat{\vv_i}^1, \ldots, \widehat{\vv_i}^K]$ concatenates $K$ feature descriptors output by the teacher models, and $h: \real^d \rightarrow \real^{d \cdot K}$ is a linear layer which maps the feature descriptor $\vv_i$ to a higher dimensional space where the cosine distance can be computed (the classifier $g\prm{W}$ is then stacked on top of $h(\vv_i)$).
We refer to this variant as \CBDens in our experiments.

The feature extractors of the teacher models account for the complementary information of the long-tailed training set. By jointly distilling knowledge from them, we transfer the enhanced feature representations to the student feature extractor $f_{\theta}$, which eases the learning of the classifier $g_W$.  
\section{Experiments}
\label{sec:exp}

\subsection{Experimental Setup}
\label{sec:setup}

\head{Datasets.} We experiment with three long-tailed datasets, namely, ImageNet-LT~\cite{OLTR}, iNaturalist18~\cite{inaturalist} and iNaturalist17\cite{inaturalist2017}.
Please refer to Section~\ref{sec:datadetails} of the appendix for details of each dataset.
Top-1 accuracy is the evaluation metric for all experiments.
We also follow the protocol in~\cite{OLTR} to report the accuracies for \emph{many-shot} classes (more than 100 images per class), \emph{mid-shot} classes (between 20 and 100 images) and \emph{few-shot} classes (less than 20 images), separately.

\head{Implementation Details.}
We use the ResNet-\{50,152\}~\cite{HZRS16} architectures for ImageNet-LT, and
ResNet-\{50,101\} for iNaturalist17 and iNaturalist18. 
See Section~\ref{sec:traindetails} of the appendix for training details.
The scaling parameter in Eq.~\eqref{eq:featdistill} is set to $\beta=100$ based on the accuracy in the ImageNet-LT validation set (see Section~\ref{sec:finetune} in Appendix). 
Other parameters, such as $\alpha$ and the number of teacher models $K$ are chosen based on the experiments in Section~\ref{sec:ablation}.

\subsection{Ablation Study}
\label{sec:ablation}

We  study the impact of some of the hyper-parameters and components of \CBD.
All experiments in this section are evaluated on the validation set of ImageNet-LT.

\begin{figure}
\centering
\input{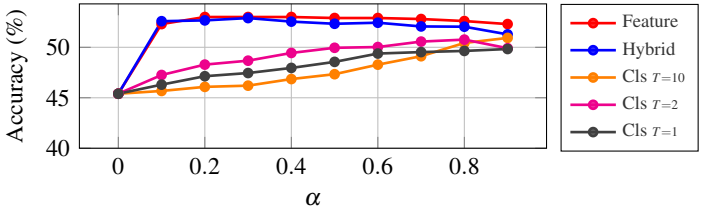}
\begin{tikzpicture}
    \tikzstyle{every node}=[font=\small]
\begin{axis}[%
	width=0.60\textwidth,
	height=0.27\textwidth,
	xlabel={$\alpha$},
  ymin = 40,
   	% xtick={0.1,0.2,0.3,0.4,0.5,0.6,0.7,0.8,0.9,1.0},
   	% xticklabels={$10^0$,$10^1$,$10^2$,$10^3$,$10^4$,$10^5$},
   	% xmode=log,
    grid=both,
	ylabel= {Accuracy (\%)},
	xminorticks=false,
  	legend cell align={left},
    legend pos=outer north east,
    title style={yshift=-1.1ex}
]

	\addplot[color=red,     solid, mark=*,  mark size=1.5, line width=1.0] table[x=alpha, y expr={\thisrow{fd}*100}] \alphavsdistillation;\leg{\scriptsize{Feature}}
	\addplot[color=blue,     solid, mark=*,  mark size=1.5, line width=1.0] table[x=alpha, y expr={\thisrow{fdcd}*100}] \alphavsdistillation;\leg{\scriptsize{Hybrid}}
  \addplot[color=orange,     solid, mark=*,  mark size=1.5, line width=1.0] table[x=alpha, y expr={\thisrow{cdten}*100}] \alphavsdistillation;\leg{\scriptsize{Cls $\scriptscriptstyle T=10$}}
	\addplot[color=magenta,     solid, mark=*,  mark size=1.5, line width=1.0] table[x=alpha, y expr={\thisrow{cdtwo}*100}] \alphavsdistillation;\leg{\scriptsize{Cls $\scriptscriptstyle T=2$}}
  \addplot[color=darkgray,     solid, mark=*,  mark size=1.5, line width=1.0] table[x=alpha, y expr={\thisrow{cdone}*100}] \alphavsdistillation;\leg{\scriptsize{Cls $\scriptscriptstyle T=1$}}

\end{axis}
\end{tikzpicture}
\caption{Impact of $\alpha$ in different distillation techniques. Experiments are conducted with ResNet-50 on the ImageNet-LT validation set.
\label{fig:distill}
}
\end{figure}

\head{Distillation.}
We first evaluate different distillation techniques, \ie feature distillation (Eq.~\eqref{eq:featdistill}) and classification distillation (Eq.~\eqref{eq:distill}) in Figure~\ref{fig:distill}.
We report classification distillation with different temperature $T$ values.
We also show the impact of the distillation coefficient $\alpha$ in the same figure.
This parameter controls the strength of distillation in the loss function, see Eq.~\eqref{eq:featdistill}.

Figure~\ref{fig:distill} shows that $T=2$ achieves the highest accuracy for classification distillation.
Feature distillation outperforms all variants of classification distillation.
It also outperforms a variant (Hybrid) which combines feature and classification distillation ($T=2$) together.
Feature distillation is also more stable for different $\alpha$.
This is expected, as the first stage model (instance sampling) produces relatively good features but a sub-optimal classifier.
Therefore, it is more beneficial to transfer information directly from the features, rather than the classifier.
It is also shown that feature distillation remains relatively stable when $\alpha >0$.
Note that $\alpha=0$ means that no distillation loss term is used during the training,
which is equivalent to class-balanced sampling.
We set $\alpha=0.4$, which gives the top performance in Figure~\ref{fig:distill}, for the remainder of our experiments.
\begin{table}
%\small
  \scriptsize 
% \begin{center}
\begin{tabular}{@{\msp}l@{\msp}|@{\msp}l@{\msp}|@{\msp}l@{\msp}|@{\msp}l@{\msp}}
\toprule
						& Vanilla		& Data Aug.		& Acc. (\%) \\
\midrule 
\multirow{2}{*}{$K=1$}  & \OK 			& - 			& 52.7 \\
						& -				& \OK			& 53.9 \\
\midrule
\multirow{3}{*}{$K=2$}  & \OK \OK 		& - 			& 54.2 \\
						&	-			& \OK \OK		& 55.5 \\
						& \OK			& \OK			& 56.2 \\
\midrule
$K=3$					& \OK			& \OK \OK		& 56.7 \\
\midrule
$K=4$					& \OK \OK		& \OK \OK 		& 56.9 \\
\midrule
$K=5$					& \OK \OK		& \OK \OK \OK	& 56.9 \\
\bottomrule
\end{tabular}
% \end{center}

  \caption{\textbf{Different ensembles of teachers.} Comprehensive evaluation of different types of $K$ teacher models on the ImageNet-LT validation set with ResNet-50. Each row corresponds to a different ensemble. Multiple \OK refer to multiple models of the same type trained with different random seeds.
  \label{tab:ensemble}}
\end{table}

\head{Number of teacher models.}
We train $K$ teacher models when ensembling is used. 
The ensemble may contain teacher models of different types, \ie \emph{standard} and \emph{data augmentation}.
When using the same type multiple times, \eg two \emph{standard} models, each model is trained with different random seeds to achieve diversity between models.
These teacher models are then fused into a single model with distillation -- Eq.~\eqref{eq:ensembledistill}.
We refer to this variant of our method as \CBDens.

Table~\ref{tab:ensemble} shows the impact of different number of \emph{standard} and \emph{data augmentation} models when used in an ensemble.
We report all combinations for $K=1$ and $K=2$, but only show the variant with the highest accuracy for $K>2$.
For $K=1$, the \emph{data augmentation} model achieves a better performance than the \emph{standard} model.
Nevertheless, we achieve the best accuracy with some combination of \emph{standard} and \emph{data augmentation} models for $K >1$.
The validation accuracy saturates after $K=4$, therefore we use the $K=4$ for \CBDens for the remainder of our experiments.

\subsection{Comparison with Baselines}
\label{sec:expbaseline}

\begin{table*}

\scriptsize 
\begin{center}
\begin{tabular}{l|lll|l}
\toprule
Method 					& Many-shot 					& Mid-shot 				& Few-shot 					& All 					\\
\midrule
% \multicolumn{9}{c}{\textbf{\textsc{Single-Stage Models}}}  \\ \midrule
Standard - Instance 	& 66.6 							& 40.4 					& 13.0 						& 46.7  				\\
Standard - Class Bal. 	& 60.4							& 40.0					& 14.3						& 44.3					\\
Data Aug. - Instance 	& 66.2 							& 38.6					& 11.2 						& 45.4  				\\
Data Aug. - Class Bal. 	& 58.4 							& 45.2					& 19.9 						& 46.8  				\\
\midrule
% \multicolumn{9}{c}{\textbf{\textsc{Two-Stage Models}}}  \\ \midrule
Standard - Fine-tuning				& 62.8							& 46.1					& 24.8						& 49.6					\\
Standard - Classifier Re-Training	& 62.9 							& 46.0 					& 25.7 						& 49.8  				\\
Data Aug. - Fine-tuning				& 63.1							& 48.4					& 26.9						& 51.1					\\
Data Aug. - Classifier Re-Training	& 62.2 							& 47.1 					& 27.8 						& 50.3  				\\
\midrule
Teacher Ensemble 					& \textbf{71.6}					& 44.4					& 13.8						& 50.7					\\
\midrule
\textbf{Ours} - \CBD 				& 65.2 							& 48.0 					& 25.9						& 51.6					\\
\textbf{Ours} - \CBDens 			& \textbf{68.5} 				& \textbf{52.7} 		& \textbf{29.2} 			& \textbf{55.6}			\\

\bottomrule
\end{tabular}
\end{center}

  \caption{\textbf{Baseline comparison.} Comprehensive evaluation on ImageNet-LT (test set) with the ResNet-50 architecture. The accuracy for many-shot 
, mid-shot 
and few-shot 
classes are reported separately. 
  \label{tab:baseline}}
\end{table*}

We compare our method against various baselines.
The results are reported on the ImageNet-LT test set. 
Please refer to Section~\ref{sec:baselines} more detailed description of each baseline.
For single-stage models, we evaluate \emph{standard} and \emph{data augmentation} models separately with instance and class balanced sampling strategies.
For two-stage models, we evaluate \emph{fine-tuning}~\footnote{The network is fine-tuned for 10 epochs with $0.01$ learning rate in the second stage, which was the best setup for this method on ImageNet-LT} and \emph{classifier-retraining}, which is our re-implementation of \emph{cRT}~\cite{kang2019decoupling} with the cosine classifier.
We also evaluate the \emph{data augmentation} version of two-stage baselines, where the first stage is trained with the \emph{data augmentation} model and the second stage is trained with the \emph{standard} model.
Finally, we evaluate the \emph{Teacher Ensemble} baseline, which simply takes the average output of teacher models during testing.

Table~\ref{tab:baseline} reports the comparisons against the baselines.
We report the accuracy of many-shot, mid-shot, and few-shot classes separately, in addition to the overall accuracy for all classes.
When compared to other two-stage models, both \CBD and \CBDens show significant improvements.
This confirms that our method is a better option as a two-stage model, even if a single teacher model is used (\CBD).
Note that the two-stage baselines reduce the accuracy of many-shot classes in the second stage.
Ensemble baselines improve the performance for many-shot classes, but show no improvements for mid-shot and few-shot classes.
This is not the case for \CBDens on ImageNet-LT, which shows improvements for \emph{all} class types.
We also observe that the \emph{data augmentation} model does not show any significant improvements except for \CBDens. 
This demonstrates that our method is capable of combining diverse models in the most effective way.

\head{Longer training of baselines.}
In order to justify that the improvement is not only due to the longer training, we train the \emph{Standard - Instance model} for two times the number of epochs.
This means that the model is trained for $180$ epochs on ImageNet-LT and $400$ epochs on iNaturalist18, \ie the total number of epochs it takes to train \CBD.
We obtain $47.1$ and $64.7$ overall accuracy for ImageNet-LT and iNaturalist18, respectively.
When compared to the \emph{Standard - Instance model} on Table~\ref{tab:baseline}, the improvement is minimal, which confirms that the improvements of \CBD are not due to longer training.

We also repeat the same procedure for the \emph{Classifier re-Training} baseline, where we train the linear model for $90$ (ImageNet-LT) and $200$ (iNaturalist18) epochs in the second stage.
We obtain $50.1$ and $67.2$ for ImageNet-LT and iNaturalist18, respectively.
When compared to the \emph{Classifier re-Training} model on Table~\ref{tab:baseline}, the gains are again minimal.
This again confirms that the efficacy of \CBD and \CBDens is not due to the longer training times.

\head{Complexity.}
\CBD requires higher training complexity compared to other baselines.
A network is trained from scratch in each stage.
We demonstrate that if other baselines (\emph{Instance}  and \emph{Classifier re-Training} ) are given the same amount of training resources, their performance is still lower than \CBD.
\CBDens requires training multiple ($K=4$) teacher models in the first stage, which further increases the training complexity.
However, the teacher models do not interact with each other during the training, which means that all teacher models can be trained in parallel, which can significantly improve the overall time for training. Memory consumption does not depend on the scale of the dataset, as it is fixed (e.g. 4 ResNet-50 models) regardless of the size of the dataset.
Note that both \CBD and \CBDens require a single model during the inference.
Therefore, the test time efficiency remains the same as for all the other baselines.

\begin{table*}
%\small
  \makebox[\textwidth][c]{
\scriptsize 
% \begin{center}
\scalebox{0.95}{
    \begin{tabular}{@{\ssp}l@{\ssp}|@{\ssp}c@{\ssp}|@{\ssp}c@{\ssp}}
    \toprule
    \multicolumn{3}{c}{ImageNet-LT}   \\
    \midrule
     Method									& R-50		    & R-152	    \\
    \midrule
    LWS~\cite{kang2019decoupling}			& 47.7			& 50.5     \\
    cRT~\cite{kang2019decoupling}			& 47.3			& 50.1     \\
    cRT+SSP~\cite{yang2020rethinking} 		& 51.3			& -     	\\
    Logit Adj.~\cite{menon2021long}         & 51.1          & 52.1      \\
    ELF(LDAM)~\cite{duggal2020elf}          & 52.0          & -          \\
    \midrule
    \textbf{Ours} - \CBD  					                     & 51.6			    & 53.9	\\
    % \textbf{Ours} - \CBD+\cite{menon2021long}       & 52.2             & -  \\
    \textbf{Ours} - \CBDens					                     & 55.6	            & \textbf{57.7}	\\
    % \textbf{Ours} - \CBDens+\cite{menon2021long}   & \textbf{56.1}    & \textbf{-} \\
    \bottomrule
    \end{tabular}
}
\hspace{1ex}
\scalebox{0.95}{
    \begin{tabular}{@{\ssp}l@{\ssp}|@{\ssp}c@{\ssp}|@{\ssp}c@{\ssp}}
    \toprule
    \multicolumn{3}{c}{iNaturalist18}   \\
    \midrule
     Method                                 & R-50          & R-101     \\
    \midrule
    LWS~\cite{kang2019decoupling}           & 69.5          & 69.7     \\
    cRT~\cite{kang2019decoupling}           & 68.2          & 70.7     \\
    cRT+SSP~\cite{yang2020rethinking}       & 68.1          & -         \\
    Logit Adj.~\cite{menon2021long}         & 68.4          & 70.8      \\
    ELF(LDAM)~\cite{duggal2020elf}          & 69.8          & -          \\
    \midrule
    \textbf{Ours} - \CBD                                            & 68.4              & 70.5  \\
    % \textbf{Ours} - \CBD+\cite{menon2021long}          & 69.4              & -  \\   
    \textbf{Ours} - \CBDens                                         & \textbf{73.6}     & \textbf{75.3} \\
    % \textbf{Ours} - \CBDens+\cite{menon2021long}      & \textbf{73.6}     & \textbf{-} \\
    \bottomrule
    \end{tabular}
}
\hspace{1ex}
\scalebox{0.95}{
    \begin{tabular}{@{\ssp}l@{\ssp}|@{\ssp}c@{\ssp}|@{\ssp}c@{\ssp}}
    \toprule
    \multicolumn{3}{c}{iNaturalist17}   \\
    \midrule
     Method                                 & R-50          & R-101     \\
    \midrule
     CB~\cite{CBLoss}                       & 58.1          & 60.9     \\
     Rethinking CB~\cite{Jamal_2020_CVPR}   & 59.4          & -                 \\
     Feature Aug.~\cite{chu2020feature}     & 62.0          & 65.9              \\
     cRT~\cite{kang2019decoupling}          & 63.9          & 65.2              \\
     BBN~\cite{Zhou_2020_CVPR}              & 65.8          & -                 \\
    \midrule
    \textbf{Ours} - \CBD                                            & 64.6                  & 66.5              \\
    % \textbf{Ours} - \CBD+~\cite{menon2021long}          & 62.9                  & -  \\       
    \textbf{Ours} - \CBDens                                         & \textbf{69.3}         & \textbf{71.3}     \\
    % \textbf{Ours} - \CBDens+~\cite{menon2021long}      & 66.5                  & \textbf{-} \\    
    \bottomrule
    \end{tabular}
}
}

  \caption{\textbf{State-of-the-art comparison.} Comparison of \CBD variants against the state of the art with ResNet-50 and ResNet-152.
  \label{tab:sota}}
\end{table*}

\subsection{Comparison with State of the Art}

Table~\ref{tab:sota} compares \CBD and \CBDens with $K=4$ teachers to the state of the art on ImageNet-LT, iNaturalist18 and iNaturalist17 datasets, respectively.
Our method shows consistent improvement for all datasets with different network architectures.
On ImageNet-LT, we observe $3.6\%$ improvement with \CBDens (ResNet-50) over the prior best.
\CBDens outperforms the state of the art on iNaturalist18 (iNaturalist17) by $3.8\%$ ($3.5\%$) with ResNet-50.
Relative improvement is even higher when a larger network is used;
we observe $5.5\%$ improvement over state of the art with \CBDens with ResNet-152 in ImageNet-LT, and $4.5\%$ improvement over state of the art in iNaturalist18 with ResNet101.
See Section~\ref{sec:compsota} of the Appendix for result for each class split separately. 

To investigate the compatibility of \CBD with existing methods, we also include a variant where the loss function in \CBD is replaced by the loss function proposed in the work of Menon \etal ~\cite{menon2021long}. On ImageNet-LT, 
\CBD + Logit Adjustment~\cite{menon2021long} gains $0.6\%$ over \CBD, i.e., it obtains $52.2$ accuracy, and \CBDens +~\cite{menon2021long} improves $0.5\%$ over \CBDens, i.e., it achieves $56.1$ accuracy.

%conclusions
\section{Conclusions}
In this paper, we have introduced a new two-stage method for long-tailed recognition called \CBD.
Our approach leverages knowledge distillation to combine information from two sampling strategies. 
Both the feature representation and the classifier evolve between stages, leading to a more effective model.
We thoroughly evaluate the effectiveness of our method by comparing it against baselines and previous work.
Our experiments demonstrate that \CBD significantly improves the state of the art in long-tailed recognition benchmarks.

\clearpage
\bibliography{egbib}
\newpage

\appendix
\section{Appendix}

\subsection{Dataset details}
\label{sec:datadetails}
We use three long-tailed datasets in our experiments, namely, ImageNet-LT~\cite{OLTR}, iNaturalist18~\cite{inaturalist} and iNaturalist17\cite{inaturalist2017}.
ImageNet-LT is an artificially created subset of the original ImageNet dataset~\cite{ImageNet} where the classes follow a long-tailed distribution.
It has $1000$ classes and the number of training images per class varies from $5$ to $1280$.
iNaturalist17 and iNaturalist18 training sets are long-tailed by nature. 
iNaturalist17 contains $5,089$ classes with the range of $9$ to $3919$ images per class.
iNaturalist18 contains $8,142$ classes with the range of $2$ to $1000$ images per class. 
The validation and test sets for ImageNet-LT and iNaturalist18  are balanced. 
The validation set of iNaturalist17 set is more balanced than the training set.

\subsection{Training details}
\label{sec:traindetails}
Throughout our experiments, the networks are trained for $90$ epochs on ImageNet-LT and $200$ epochs on iNaturalist17 and iNaturalist18 in both stages, to make the results comparable with the existing work.
We also report the performance with more epochs in Section~\ref{sec:expbaseline}.
When training the model, we use a batch size of $256$, learning rate of $0.2$ which decays to $0$ with cosine learning rate schedule~\cite{loshchilov2016sgdr}, and SGD optimizer with momentum $0.9$.

\begin{figure}
\centering
\input{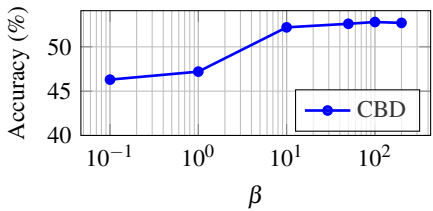}
\begin{tikzpicture}
    \tikzstyle{every node}=[font=\small]
\begin{axis}[%
	width=0.48\textwidth,
	height=0.25\textwidth,
	xlabel={$\beta$},
  ymin = 40,
   	xmode=log,
    grid=both,
	ylabel= {Accuracy (\%)},
	xminorticks=false,
  	legend cell align={left},
  	legend pos=south east,
]

	\addplot[color=blue,     solid, mark=*,  mark size=1.5, line width=1.0] table[x=beta, y expr={\thisrow{fd}*100}] \betavsacc;\leg{\CBD}
\end{axis}
\end{tikzpicture}
\caption{Impact of $\beta$ scaling hyperparameter for \CBD. Experiments are conducted with ResNet-50 on ImageNet-LT validation set with $\alpha =0.4$.
\label{fig:beta}
}
\end{figure}

\subsection{Impact of $\beta$}
\label{sec:finetune}
We first demonstrate the impact of $\beta$ in Equation~\eqref{eq:featdistill}.
This parameter scales the feature distillation loss in the loss objective.
Figure~\ref{fig:beta} shows the accuracy for the ImageNet-LT validation set with different $\beta$.
We see that the accuracy is relatively stable for $\beta \geq 10$.
We set $\beta =100$ for all of our experiments as this gives the highest accuracy for the validation set.

\begin{table*}
%\small
  \scriptsize	
\begin{center}
\begin{tabular}{l|lll|l}
\toprule
						\multicolumn{4}{c}{iNaturalist18} \\
\midrule 
Method 					& Many-shot 					& Mid-shot 				& Few-shot 					& All				\\
\midrule
% \multicolumn{9}{c}{\textbf{\textsc{Single-Stage Models}}}  \\ \midrule
Standard - Instance 	& 76.4					& 66.2 					& 57.9 						& 64.0  			\\
Standard - Class Bal. 	& 59.3							& 65.2					& 62.6						& 63.6				\\
Data Aug. - Instance 	& 74.5 							& 62.5 					& 52.5 						& 59.8  			\\
Data Aug. - Class Bal. 	& 55.9 							& 65.1 					& 62.6 						& 63.1  			\\
\midrule
% \multicolumn{9}{c}{\textbf{\textsc{Two-Stage Models}}}  \\ \midrule
Standard - Fine-tuning				& 69.6							& 68.5					& 66.1						& 67.6				\\
Standard - Classifier Re-Training	& 74.1 							& 68.1 					& 64.2 						& 67.2				\\
Data Aug. - Fine-tuning				& 69.5							& 68.1					& 65.3						& 67.1				\\
Data Aug. - Classifier Re-Training	& 69.9 							& 65.7 					& 62.9 						& 65.1				\\
\midrule
Teacher Ensemble 					& \textbf{81.9}					& 71.9					& 63.6						& 69.7				\\
\midrule
% \multicolumn{9}{c}{\textbf{\textsc{Ours}}}  \\ \midrule
\textbf{Ours} - \CBD 				& 70.5 							& 69.5 					& 66.5 						& 68.4	 			\\
\textbf{Ours} - \CBDens 			& 75.9 							& \textbf{74.7} 		& \textbf{71.5} 			& \textbf{73.6}		\\

\bottomrule
\end{tabular}
\end{center}

  \caption{\textbf{Baseline comparison.} Comprehensive evaluation on iNaturalist18 (validation set) with the ResNet-50 architecture. The accuracy for many-shot 
, mid-shot 
and few-shot 
classes are reported separately. 
  \label{tab:baselineinat}}
\end{table*}

\subsection{Baseline comparison on iNaturalist18}
Table~\ref{tab:baselineinat} reports the comparisons against the baselines on iNaturalist18.
Similar to Table~\ref{tab:baseline}, we report the accuracy of many-shot, mid-shot, and few-shot classes separately, in addition to the overall accuracy for all classes.
Our conclusions are similar to the baseline comparison on ImageNet-LT.
When compared to other two-stage models, both \CBD and \CBDens show significant improvements.
\emph{Teacher Ensemble} improves the accuracy of many-shot classes at the expense of mid-shot and few-shot classes.
This confirms that our method is a better option as a two-stage or ensemble model.

\begin{table}
%\small
  \footnotesize
\begin{center}
\begin{tabular}{l|l|l}
\toprule
						& ImageNet-LT	& iNaturalist18 \\
\midrule 
Instance - NCM			& 49.0			& 62.8 	\\
Fine-tuning - NCM		& 48.8			& 64.1	\\
\midrule 
\CBD - NCM				& 50.7			& 64.5 \\
\CBDens - NCM			& 54.0			& 69.2 \\
\bottomrule
\end{tabular}
\end{center}

  \caption{\textbf{Evaluation with NCM.} Classification accuracy with the non-parametric Nearest Centroid Mean~\cite{kang2019decoupling} classifier. ResNet-50 architecture is used for both datasets.
  \label{tab:ncm_table}}
\end{table}

\begin{figure*}
\small
\footnotesize
{\setlength{\fboxsep}{2.0pt}
\centering
\begin{tabular}{@{\ssp}c@{\ssp}c@{\ssp}c@{\ssp}c@{\ssp}c@{\ssp}c@{\ssp}}
\includegraphics[width=42pt,height=42pt]{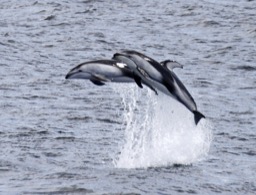}&
\fcolorbox{green}{green}{\includegraphics[width=42pt,height=42pt]{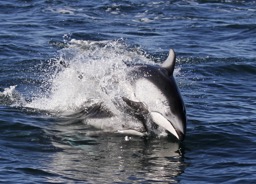}}&
\fcolorbox{green}{green}{\includegraphics[width=42pt,height=42pt]{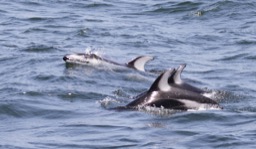}}&
\fcolorbox{green}{green}{\includegraphics[width=42pt,height=42pt]{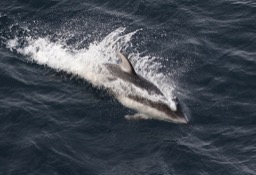}}&
\fcolorbox{green}{green}{\includegraphics[width=42pt,height=42pt]{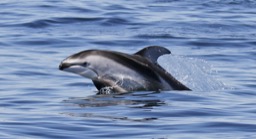}}&
\fcolorbox{green}{green}{\includegraphics[width=42pt,height=42pt]{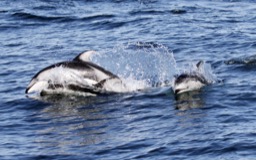}}\\[0pt]
white-sided dolphin (19)& \footnotesize{\CBD Rank 1}& \footnotesize{\CBD Rank 2}& \footnotesize{\CBD Rank 3}& \footnotesize{\CBD Rank 4}& \footnotesize{\CBD Rank 5}\\
% \includegraphics[width=70pt,height=70pt]{fig/examples_places/class39_cafeteria/top_499}&
% &&(1, 1.00)&(2, 1.00)&(4, 1.00)&(5, 1.00)&(50, 1.00)&(100, 0.98)&(500, 0.91)\\ & &
[5pt]
&
\fcolorbox{red}{red}{\includegraphics[width=42pt,height=42pt]{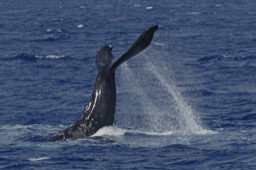}}&
\fcolorbox{red}{red}{\includegraphics[width=42pt,height=42pt]{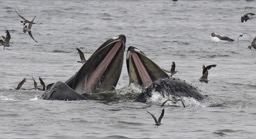}}&
\fcolorbox{red}{red}{\includegraphics[width=42pt,height=42pt]{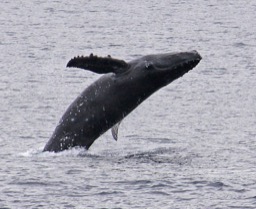}}&
\fcolorbox{red}{red}{\includegraphics[width=42pt,height=42pt]{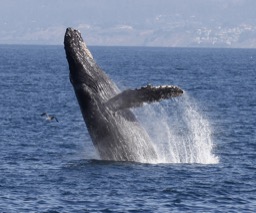}}&
\fcolorbox{red}{red}{\includegraphics[width=42pt,height=42pt]{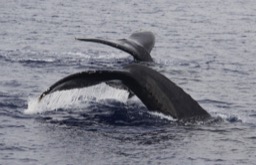}}\\[0pt]
& \footnotesize{\emph{Instance} Rank 1}& \footnotesize{\emph{Instance} Rank 2}& \footnotesize{\emph{Instance} Rank 3}& \footnotesize{\emph{Instance} Rank 4}& \footnotesize{\emph{Instance} Rank 5}\\
[10pt]
\includegraphics[width=42pt,height=42pt]{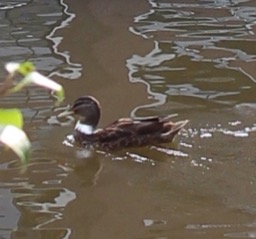}&
\fcolorbox{green}{green}{\includegraphics[width=42pt,height=42pt]{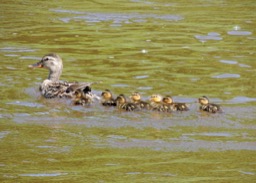}}&
\fcolorbox{green}{green}{\includegraphics[width=42pt,height=42pt]{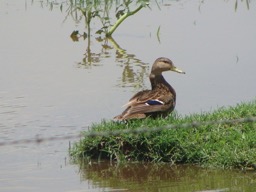}}&
\fcolorbox{red}{red}{\includegraphics[width=42pt,height=42pt]{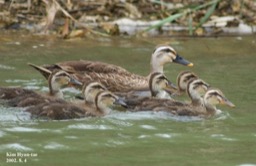}}&
\fcolorbox{green}{green}{\includegraphics[width=42pt,height=42pt]{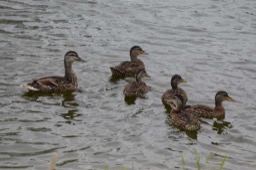}}&
\fcolorbox{green}{green}{\includegraphics[width=42pt,height=42pt]{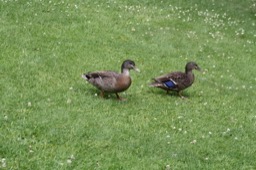}}\\[0pt]
mallard (1000)& \footnotesize{\CBD Rank 1}& \footnotesize{\CBD Rank 2}& \footnotesize{\CBD Rank 3}& \footnotesize{\CBD Rank 4}& \footnotesize{\CBD Rank 5}\\
% \includegraphics[width=70pt,height=70pt]{fig/examples_places/class39_cafeteria/top_499}&
% &&(1, 1.00)&(2, 1.00)&(4, 1.00)&(5, 1.00)&(50, 1.00)&(100, 0.98)&(500, 0.91)\\ & &
[5pt]
&
\fcolorbox{red}{red}{\includegraphics[width=42pt,height=42pt]{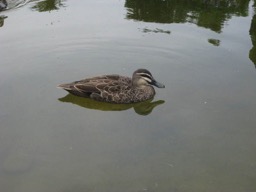}}&
\fcolorbox{green}{green}{\includegraphics[width=42pt,height=42pt]{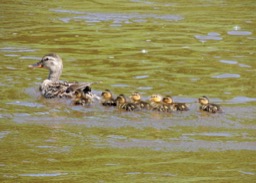}}&
\fcolorbox{red}{red}{\includegraphics[width=42pt,height=42pt]{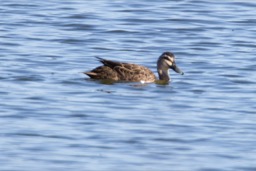}}&
\fcolorbox{red}{red}{\includegraphics[width=42pt,height=42pt]{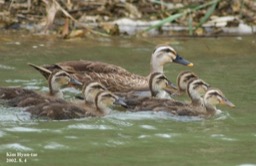}}&
\fcolorbox{red}{red}{\includegraphics[width=42pt,height=42pt]{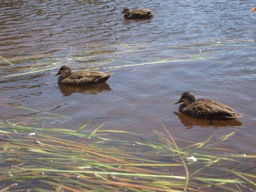}}\\[0pt]
& \footnotesize{\emph{Instance} Rank 1}& \footnotesize{\emph{Instance} Rank 2}& \footnotesize{\emph{Instance} Rank 3}& \footnotesize{\emph{Instance} Rank 4}& \footnotesize{\emph{Instance} Rank 5}\\
[13pt]
\end{tabular}
}

\caption{\textbf{Qualitative results.} Test images from the iNaturalist18 dataset are depicted on the left, along with their class labels and associated number of training images. For each test image, we show its nearest neighbors from the training set using the feature representations of \CBD (top) and \emph{Instance sampling} (bottom). Green and red boundaries denote training images from the same or different classes, respectively.
\label{fig:examples}
}
\end{figure*}

\subsection{Quality of the feature representation}
We now investigate if our method indeed improves the feature representation.
To this end, we measure the accuracy with a non-parametric classifier, \ie \emph{NCM}~\cite{guerriero2018deepncm,kang2019decoupling}.
NCM computes the classification vectors for each class by taking the mean of all vectors belonging to that class.
Thus, the classification accuracy is directly related to the feature representation quality.
Table~\ref{tab:ncm_table} shows the classification performance with the NCM classifier.
We test our method against \emph{Instance}, which is the same feature representation used by cRT, and also \emph{Fine-tuning}, which also updates the feature representation in the second stage.
It is shown that \CBD and \CBDens achieves higher accuracy, which validates our claims that the feature representation is also improved.

Figure~\ref{fig:examples} presents qualitative examples that demonstrate the quality of the feature representation produced by \CBD, in comparison to the features of \emph{Instance}, \ie, the same representation used in \emph{cRT} method.
For each test image, the top-5 nearest neighbor training images are depicted.
\CBD improves the challenging cases by learning a much improved feature representation for tail classes (\emph{white-sided dolphin}).
Features of dolphins and whales seem to be collapsed together with the feature representation used in \emph{Instance}.
Our method obtains improvements even for head classes, such as \emph{mallard}, where our learned feature can more easily distinguish similar species based on very detailed information.

\subsection{Upper-bound performance}
Even though our method brings significant improvements over other baselines, we investigate whether there is further room for improvement.
To illustrate the headroom for improvement in terms of feature learning, we propose an experiment where we assume to have the \emph{optimal} feature representation for a long-tailed recognition problem.
First, we train a ResNet-50 model $\phi^*\prm{\theta,W}$ on the full ImageNet~\cite{ImageNet} dataset.
We remove the classifier from this model, only keeping the feature extractor $f^*\prm{\theta}$.
This feature extractor produces the optimal feature representation for the ImageNet-LT~\cite{OLTR} dataset, in the sense that
ImageNet-LT is a subset of ImageNet, where the classes follow a long-tailed distribution.
By training $f^*\prm{\theta}$ on the \emph{full} ImageNet, we essentially learn the best possible feature extractor for ImageNet-LT for a given architecture.
We now fix $f^*\prm{\theta}$ and only train the classifier $g\prm{W}$ on ImageNet-LT with class-balanced sampling.
The resulting model achieves $73.5\%$ top-1 accuracy, which can be seen as an upper bound on ImageNet-LT with ResNet-50.
On the other hand, \CBDens achieves $55.6\%$ top-1 accuracy.
This suggests that there is still a lot of room for improvement on long-tailed datasets.

\subsection{Comprehensive comparison with State of the Art}
\label{sec:compsota}

\begin{table*}
%\small
  \scriptsize	
\begin{center}
\begin{tabular}{l|ccc|c|ccc|c}
\toprule
 Method 								& \multicolumn{4}{c}{ResNet-50}	& \multicolumn{4}{c}{ResNet-152} \\
\midrule
										& Many-shot		& Mid-shot		& Few-shot		& All					& Many-shot		& Mid-shot		& Few-shot	& All	\\
\midrule
% \multicolumn{9}{c}{\textbf{\textsc{Previous Work}}}  																								\\ 
% \midrule
 LWS~\cite{kang2019decoupling}  		& 57.1 			& 45.2			& 29.3			& 47.7 					& 60.6 			& 47.8			& 31.4		& 50.5	\\
 cRT~\cite{kang2019decoupling} 			& 58.8			& 44.0			& 26.1 			& 47.3 					& 61.8			& 46.8			& 28.4 		& 50.1  \\
 cRT+SSP~\cite{yang2020rethinking}		& -				& -				& -				& 51.3					& -				& -				& -			& -		\\
 Logit Adj.~\cite{menon2021long}		& -				& -				& -				& 51.1					& -				& -				& -			& 52.1	\\
 ELF(CE)~\cite{duggal2020elf}			& 60.7			& 45.5			& 27.7			& 48.9					& -				& -				& -			& -		\\
 ELF(LDAM)~\cite{duggal2020elf}			& 64.3			& 47.9			& 31.4			& 52.0					& -				& -				& -			& -		\\
\midrule
% \multicolumn{9}{c}{\textbf{\textsc{Ours}}}  \\ \midrule
\textbf{Ours} - \CBD 									& 65.2 			& 48.0 			& 25.9			& 51.6					& 68.1 			& 50.1			& 27.1			& 53.9  \\
\textbf{Ours} - \CBDens 								& \textbf{68.5} & \textbf{52.7} & \textbf{29.2} & \textbf{55.6}			& \textbf{71.2} & \textbf{54.5}	& \textbf{30.7}	& \textbf{57.7}  \\
\bottomrule
\end{tabular}
\end{center}

  \caption{\textbf{ImageNet-LT state-of-the-art comparison.} Comparison of \CBD variants against the state of the art with ResNet-50 and ResNet-152.
  \label{tab:sotaImnet}}
\end{table*}

\begin{table*}
%\small
  \scriptsize	
\begin{center}
\begin{tabular}{l|ccc|c|ccc|c}
\toprule
 Method 								& \multicolumn{4}{c}{ResNet-50}	& \multicolumn{4}{c}{ResNet-101} \\
\midrule
										& Many-shot		& Mid-shot		& Few-shot		& All					& Many-shot		& Mid-shot		& Few-shot	& All	\\
\midrule
% \multicolumn{9}{c}{\textbf{\textsc{Previous Work}}}  																								\\ 
% \midrule
 LWS~\cite{kang2019decoupling}  		& 71.0 			& 69.8			& 68.8			& 69.5 					& 73.9 			& 70.4			& 67.8		& 69.7	\\
 cRT~\cite{kang2019decoupling} 			& 73.2 			& 68.8			& 66.1			& 68.2 					& 71.5			& 71.3			& 69.7 		& 70.7  \\
 cRT+SSP~\cite{yang2020rethinking}		& -				& -				& -				& 68.1					& -				& -				& -			& -		\\
 Logit Adj.~\cite{menon2021long}		& -				& -				& -				& 68.4					& -				& -				& -			& 70.8	\\
 ELF(CE)~\cite{duggal2020elf}			& 67.4			& 66.3			& 65.1			& 66.0					& -				& -				& -			& -		\\
 ELF(LDAM)~\cite{duggal2020elf}			& 72.7			& 70.4			& 68.3			& 69.8					& -				& -				& -			& -		\\
 BBN~\cite{Zhou_2020_CVPR}				& -				& -				& -				& 69.6					& -				& -				& - 		& -		\\
\midrule
% \multicolumn{9}{c}{\textbf{\textsc{Ours}}}  \\ \midrule
\textbf{Ours} - \CBD 									& 70.5 			& 69.5 			& 66.5 			& 68.4					& 74.2			& 71.5			& 68.3			& 70.5  \\
\textbf{Ours} - \CBDens 								& \textbf{75.9} & \textbf{74.7} & \textbf{71.5} & \textbf{73.6}			& \textbf{77.9} & \textbf{76.5}	& \textbf{73.2}	& \textbf{75.3}  \\
\bottomrule
\end{tabular}
\end{center}

  \caption{\textbf{iNaturalist18 state-of-the-art comparison.} Comparison of \CBD variants against the state of the art with  ResNet-50 and ResNet-101.
  \label{tab:sotaInat}}
\end{table*}

\begin{table}
%\small
  \scriptsize	
\begin{center}
\begin{tabular}{l|ccc|c}
\toprule
\multicolumn{5}{c}{\textbf{ResNet-50}} \\
\midrule
										& Many-shot		& Mid-shot		& Few-shot		& All \\
\midrule
% \multicolumn{9}{c}{\textbf{\textsc{Previous Work}}}  																								\\ 
% \midrule
 Standard Instance										& 74.2			& 55.8			& 42.9			& 62.5 \\
 Classifier Re-training 								& 71.7 			& 59.3			& 53.7			& 63.9 \\
\midrule
% \multicolumn{9}{c}{\textbf{\textsc{Ours}}}  \\ \midrule
\textbf{Ours} - \CBD 									& 70.8 			& 61.0 			& 56.0 			& 64.6 \\
\textbf{Ours} - \CBDens 								& \textbf{74.3} & \textbf{66.4} & \textbf{62.0} & \textbf{69.3}	\\
\midrule
\multicolumn{5}{c}{\textbf{ResNet-101}} \\
\midrule
										& Many-shot		& Mid-shot		& Few-shot		& All \\
\midrule
% \multicolumn{9}{c}{\textbf{\textsc{Previous Work}}}  																								\\ 
% \midrule
 Standard Instance										& 75.2			& 57.7			& 45.7			& 64.1	\\
 Classifier Re-training 								& 72.9			& 60.6			& 54.9 			& 65.2  \\
\midrule
% \multicolumn{9}{c}{\textbf{\textsc{Ours}}}  \\ \midrule
\textbf{Ours} - \CBD 									& 73.3			& 62.5			& 57.9			& 66.5 \\
\textbf{Ours} - \CBDens 								& \textbf{76.8} & \textbf{68.1}	& \textbf{63.2}	& \textbf{71.3}	\\
\bottomrule
\end{tabular}

\end{center}

  \caption{\textbf{iNaturalist17 comprehensive comparison.} Comparison of \CBD variants against the other methods with ResNet-50 and ResNet-101. The accuracy for many-shot ($>100$ images), mid-shot ($20$-$100$ images) and few-shot ($<20$ images) classes are reported separately.
  \label{tab:sotaInat17comp}}
\end{table}

Table~\ref{tab:sota} shows the overall accuracy for the ImageNet-LT, iNaturalist18, and iNaturalist17 datasets.
We now show results separately for \textit{low-shot}, \textit{mid-shot} and \textit{many-shot} classes on Tables~\ref{tab:sotaImnet},~\ref{tab:sotaInat} and~\ref{tab:sotaInat17comp}.
Note that the other methods~\cite{chu2020feature, CBLoss, Jamal_2020_CVPR,  Zhou_2020_CVPR} do not report such detailed results on iNaturalist17, thus we could not include them in the comparison.
Thus we compare our method with emph{Standard Instance}, our implementation of classifier re-training~\cite{kang2019decoupling}.
Our method also outperforms the prior work for each class split (many-shot, mid-shot, few-shot) in all scenarios.
We do not sacrifice the accuracy of  \emph{many-shot} classes to increase the overall accuracy;
we achieve a higher overall accuracy by improving \emph{many-shot} accuracy as well as the accuracy of other groups.

\end{document}